\documentclass{article}

\usepackage{arxiv}

\usepackage[utf8]{inputenc} % allow utf-8 input
\usepackage[T1]{fontenc}    % use 8-bit T1 fonts
\usepackage{hyperref}       % hyperlinks
\usepackage{url}            % simple URL typesetting
\usepackage{booktabs}       % professional-quality tables
\usepackage{amsfonts}       % blackboard math symbols
\usepackage{nicefrac}       % compact symbols for 1/2, etc.
\usepackage{microtype}      % microtypography
\usepackage{lipsum}

\usepackage{multicol}

\usepackage{graphicx}

\title{A Comparison of Deep Learning Performances with Other Machine Learning Algorithms on Credit Scoring Unbalanced Data}

\author{
  Louis Marceau\\
  National Bank Of Canada\\
  600 de la Gauchetière\\
  Montréal, Québec \\
  \texttt{louis.marceau@bnc.ca} \\
   \And
 Lingling Qiu \\
  National Bank Of Canada\\
  600 de la Gauchetière\\
  Montréal, Québec \\
  \texttt{lingling.qiu@bnc.ca} \\
    \And
 Nick Vandewiele \\
  National Bank Of Canada\\
  600 de la Gauchetière\\
  Montréal, Québec \\
  \texttt{nick.vandewiele@bnc.ca} \\
   \And
 Eric Charton \\
  National Bank Of Canada\\
  600 de la Gauchetière\\
  Montréal, Québec \\
  \texttt{eric.charton@bnc.ca} \\
}

\begin{document}
\maketitle

\begin{abstract}
Training models on highly unbalanced data is admitted to be a challenging task for machine learning algorithms. Current studies on deep learning mainly focus on data sets with balanced class labels or unbalanced data, but with massive amount of samples available, like in speech recognition. However, the capacities of deep learning on imbalanced data with little samples is not deeply investigated in literature, while it is a very common application context in numerous industries. To contribute to fill this gap, this paper compares the performances of several popular machine learning algorithms previously applied with success to unbalanced data set with deep learning algorithms. We conduct those experiments on a highly unbalanced data set, used for credit scoring. We evaluate various configuration including neural network optimisation techniques and try to determine their capacities when they operate with imbalanced corpora.
\end{abstract}

% keywords can be removed

\section{Introduction}

% rewrite
Forecasting a score of credit is a key aspect of risk management in financial institutions.  The aim of credit scoring is essentially to classify credit product (like credit card or loans) applicants into two classes: good payers (i.e., those who are likely to keep up with their repayments) and bad payers (i.e., those who are likely to default on their credit card loan). 

The particularity of this scoring task is that classes are highly unbalanced as bad behavior represent usually a small percentage of a customer base in a banking context. Training models on highly unbalanced data is admitted to be a challenging task for machine learning (ML) algorithms. A wide range of classification techniques have already been proposed in the credit scoring literature to tackle the specific aspect of unbalanced credit risk data, including statistical techniques, such as linear discriminant analysis, logistic regression, and non-parametric models, such nearest neighbour and decision trees. 

Deep learning (DL) ML techniques have become increasingly popular in both academic and industrial areas in the past years, as it is common to show that modern implementations of DL algorithms outperforms non-DL algorithms in numerous existing and well documented tasks. That being said, it is currently unclear from the literature if DL techniques performs well on unbalanced data set, and specifically on credit scoring data. Various domains including pattern recognition, computer vision, and natural language processing have witnessed the performances of deep neural networks. However, current studies on DL mainly focus on data sets with balanced class labels, or unbalanced data sets with massive amount of samples available. Its performance on more classical imbalanced data sets is not widely examined. 

In this paper, we conduct a compared experiment of traditional ML and DL algorithms on a highly unbalanced data set representing credit behavior. We use a data set coming from a 24 months credit card user data. We use various ML algorithms, including DL ones, to forecast the user behavior as good or bad in 12 months.   

This paper is organised as follows: in section \ref{sec:related} we investigate related work in the specific domain of classification tasks applied to unbalanced data. In section \ref{sec:context}, we explain the context of this study, risk management in finance, and more specifically the domain of credit scoring. We explain how we built our experimental corpus, and what features we extracted from it. In section \ref{sec:experiments} we conduct our experiments in two phases: first we establish what is the best ML classifier on each of our experimental corpus, then we apply DL algorithms. We used those two set of experiments to compare the performances of the two ML families of algorithms. After a discussion on the results, we finally conclude in section \ref{sec:conclusions}.    

\section{Related work}
\label{sec:related}

In their review, \cite{haixiang2017learning} of methods and applications for learning on unbalanced data, authors cite as domains extensively explored the topic Taxonomy, Chemical engineering, financial management, information technology, energy management, security management. Most of the related experimentation tend to show that since years, decision trees are the best performing ML algorithms on unbalanced data sets.

In \cite{cieslak01}, authors outline the performance of several popular decision tree splitting criteria – information gain, Gini measure, and DKM – can be used to form decision trees, and improve performances of tree construction method applied to unbalanced data.

Authors of \cite{almeida2014machine} discuss the use of 108 different classification models to determine the most fitting model, capable of dealing with the imbalanced data issue coming from a biomedical literature corpus and achieve the most satisfying results with a LMT decision tree classifier previously used with success in a unbalanced multi-label classification task related to taxonomy by \cite{charton2014using}.

In the domain of credit applications, authors of \cite{brown2012experimental} compare several techniques that can be used in the analysis of imbalanced credit scoring data sets. The results from this empirical study indicate that the random forest and gradient boosting classifiers perform very well in a credit scoring context and are able to cope comparatively well with pronounced class imbalances in these data sets. 

Various domains including pattern recognition, computer vision, and natural language processing have witnessed the great power of deep neural networks. However, current studies on DL mainly focus on data sets with balanced class labels, while its performance on imbalanced data is not well examined.

\section{Context of this study}
\label{sec:context}

Risk management in finance and banking industry includes determining in diverse forms how risk can evolve for a given use case. Forecasting the evolution of various risk related indicators is a key activity, previously handled through rule based or regression models, and more recently with ML generated classification models. As well as using traditional classification techniques such as logistic regression, neural networks and decision trees, gradient boosting, support vector machines and random forests have been tested in the past on various aspect of client default prediction. 

Our objective in this experiment was to forecast, for a given credit card account, a probability of default after 12 months. We used for this a set of credit card account holder characteristics along with their historical credit behavior to train a ML system and see how this model could forecast \textit{good} and  \textit{bad} labels.  

In a credit scoring context,
%, imbalanced data sets are common since the number of defaulting loans or credit card in a portfolio always corresponds to a small percentage of the number of observations that do not default. This is a difficulty, specific to the domain, and have as a consequence that 
the imbalance issue can interfere directly on the classifier performance, which is biased by the majority class. 
%% rewrite
Because the majority class (the good labeled samples) is more heavily represented in the data set than the minority class (the bad labeled samples), it tends to have more influence under uncertainty cases, since the class distribution can influence learning criteria. In addition, according to \cite{weiss2001effect}, a classifier presents a lower error rate when classifying an instance belonging to the majority class, since it will have learned more information from the examples of the majority class, compared to the information learned in fewer examples from the minority class.

\begin{figure}
  \centering
      \includegraphics[width=1\textwidth]{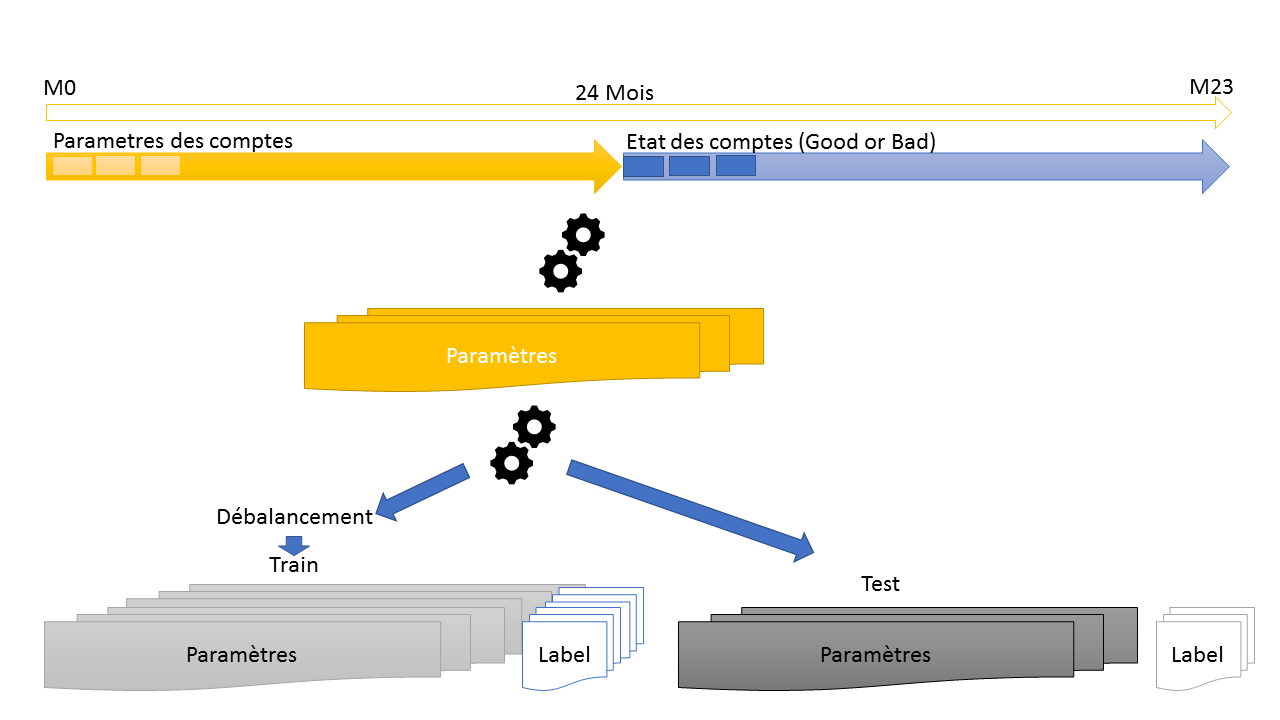}
  \caption{Corpus construction.}
 \label{fig:fig1}
\end{figure}

\subsection{Data set}

We used as a data set of two years of historical data randomly extracted from our organisation accounts data between 2015 and 2018. The extracted set is made of 1.4M accounts. This data set includes samples built from credit status, customer information and transactional data. Each sample is labelled with the value Good or Bad. We defined as a Bad label for this experiment, an account with 60 days past due, 12 months after the time the data are collected (see figure~\ref{fig:fig1}). On this data set, we kept 288,558 accounts representation (20\%) used as test corpus. 

We used the remaining data to build the train corpora. Downsizing and re-balancing can have various impact on ML applied to unbalanced data. For instance, a strong re-balancing, as it reduce substantially the amount of samples available to train the ML algorithm, can have a negative effect on the precision of the model trained. As there is no clear theoretical proposition to estimate by calculation the most adapted re-balancing amplitude we built two train corpora.  We built two down-sampling configurations for train (DS1 and DS2) as shown in table~\ref{tab:tablecorpus}.

\begin{itemize}
    \item DS1 : a corpus with a simple downsizing (the smaller class is 7.2\% of the total). 
    \item DS2 : a corpus with a strong re-balancing (the smaller class is 35\% of the total).
\end{itemize}

The test corpus reflect the exact proportion as Bad as the original corpus as show in last column of table \ref{tab:tablecorpus}.

\begin{table}[!h]
 \caption{Corpus}
  \centering
  \begin{tabular}{llll}
    \toprule
                   \\
    %\cmidrule(r){1-2}
    Corpus     & Good Samples    & Bad Samples  & smaller class\\
    \midrule
    Total   & 1442787 &  67896 & 4.38\%    \\
    Train DS1     & 700 000  & 54317  & 7.2\%    \\
    Train DS2     & 100 000 & 54317  &   35.2\%   \\
    Test          & 288 558 &  13579 &  4.38 \%   \\

    \bottomrule
  \end{tabular}
  \label{tab:tablecorpus}
\end{table}

\subsubsection{Features}

There is 66 features used by the model. Features includes account information (like credit card limit, current balance), credit status (like the delinquency cycle) and 6 categories of transactional data aggregated on the 3 months before the date of the account information collection.    

\section{Experiments}
 \label{sec:experiments}

The main purpose of our experiment is to compare ML approaches commonly applied on unbalanced data (like decision trees or SVM) with DL model. We also want to evaluate the capacities of optimization algorithms used to maximize the performance of a DL model, like AutoML, in this experimental context. Our experimental plan will consist in 2 main steps : 

\begin{enumerate}
    \item First, training common ML algorithms used for unbalanced data on our data set, and select the best configuration.
    \item Then optimising a DL algorithm on our data set from the best configuration selected.  
\end{enumerate}

We describe those two sequences of experiment in the two following subsections.

\subsection{Machine learning algorithms selection}

Our first set of experiments consist in applying 4 algorithms, Logistic Regression, SVM, Random Forest, XGBoost to the 2 corpus configurations, DS1 (re-balanced) and DS2 (downsampled). 

\subsubsection{Results}

Results of the first set of experiments are in table~\ref{tab:tableDS2} and table~\ref{tab:tableDS1}. On both experiments, with downsized corpus (DS1) and rebalanced corpus (DS2), XGBoost classifiers overperform logistic regression, SVM and Random Forest as measured by F-Score. However, the XGBoost model provides its best performances with a F-Score of 0.549 as shown in table~\ref{tab:tableDS1}. This set of experiments shows that in our experimental context, downsizing is the best solution to counteract the effects of unbalanced data on the training process. 

\begin{table}%[!h]

 \caption{Basic classifiers applied on 12 month forecast, 60 days past due, DS1 train corpus - Downsized corpus}
  \centering
  \begin{tabular}{llllll}            \\
    \toprule
    \cmidrule(r){1-2}
    Model     & Precision  & Recall & F-Score & AUC & Run time (s) \\
    \midrule
    Logistic Regression & 0.5104  & 0.5366  & 0.5232 & 0.7562 & 22.8    \\
    SVM & 0.5364 & 0.5144 & 0.5251 & 0.7467 & 149.9        \\
    Random Forest & 0.6324 & 0.4037 & 0.4928 & 0.6963 & 55.4     \\
    \textbf{XGBoost} & 0.5138  & 0.5907  &  \textbf{0.5496} & 0.7822 & 299.8    \\
    \bottomrule
  \end{tabular}
  \label{tab:tableDS1}
  
  \bigskip
 \caption{Basic classifiers applied on 12 month forecast, 60 days past due, DS2 train corpus - Re-balanced corpus}
  \centering
  \begin{tabular}{llllll}            \\
    \toprule
    \cmidrule(r){1-2}
    Model     & Precision  & Recall & F-Score & AUC & Run time (s) \\
    \midrule
    Logistic Regression & 0.298  & 0.744  & 0.426 & 0.830 & 22.8    \\
    SVM & 0.314 & 0.730 & 0.439 & 0.827 & 12.0        \\
    Random Forest & 0.308 & 0.77 & 0.44 & 0.844 & 55.4     \\
    \textbf{XGBoost} & 0.304  & 0.789  & 0.439 & 0.852    & 64.3 \\
    \bottomrule
  \end{tabular}
  \label{tab:tableDS2}

\end{table}

\subsection{Deep learning experiments}

We now apply DL algorithms and AutoML techniques on the selected corpus from the first sequence of ML experiments, the downsampled corpus DS1, and we will compare the performances of both DL and AutoML with the best classifier on DS1, XGBoost. 

\subsubsection{AutoML and Deep Learning}

%%% reecrire
When training a neural network on a data set a DL practitioner is trying to optimize and balance a neural network architecture that lends itself to the nature of the data set. After the proper architecture defined, a second step is intended to tune a set of hyper-parameters over many experiments. Typical hyper-parameters that need to be tuned include the optimizer algorithm (SGD, Adam, etc), learning rate, learning rate scheduling and regularization to name a few. Depending on the data set and problem, it can take numerous experiments to find a balance between the best neural network architecture and hyper-parameters. The complexity of having to find both an architecture and a combination of hyper-parameters is a very specific aspect of the difficulty to use DL. A difficulty that is not encountered at the same scale with ML algorithms like SVM or Decision trees.   

To answer to this problem AutoML techniques were recently proposed. Their purpose is to automatically determine a performing set of hyperparameters and a neural network architecture to train an optimized neural networks. AutoML is defined by \cite{feurer2015efficient} as \textit{the problem of automatically (without human input) producing test set predictions for a new data set within a fixed computational budget}. Numerous open frameworks are made available now to conduct AutoML experiments. We mention here Auto-Keras\footnote{\url{https://autokeras.com/}}\cite{jin2018efficient}, Auto-Weka  \footnote{\url{https://www.cs.ubc.ca/labs/beta/Projects/autoweka/}}\cite{kotthoff2017auto} and Auto-Sklearn\footnote{\url{https://github.com/automl/auto-sklearn}}\cite{feurer2015efficient}.

%%% reecrire
Both Google’s AutoML and Auto-Keras are based on the Neural Architecture Search (NAS) technique \cite{zoph2016neural}\cite{zoph2018learning}\footnote{There is an implementation difference between Google's AutoML implementation according to \cite{zoph2018learning} is based on TensorFlow framework while Auto-Keras is based on Pytorch framework. There is also some marginal algorithmic differences as Auto-Keras authors introduces various heuristics to improve the performances of their system.}. We observe that none of the proposed AutoML frameworks were tested in their respective descriptive papers on unbalanced data sets like ours (\cite{zoph2016neural} is tested on an image recognition framework and experiment in  \cite{jin2018efficient} for Auto-Keras are based on MNIST, CIFAR and FASHION, 3 images recognition data sets). 

Considering the specific nature and difficulty of comparing traditional ML algorithmic approaches with DL approaches, we decided to include in our experimental plan an AutoML training, using Auto-Keras. We considered important to do so, as, according to the plasticity of any DL model configuration, one could legitimately claim that we did not found the proper neural network configuration or architecture in our experiments, making it difficult to compare with other algorithms used. We considered AutoML output as an acceptable experiment to potentially refute this legitimate claim.

As of today, we did not found in literature previous application of NAS technique to optimize a DL model and compare it to non DL models on unbalanced data set like ours.     

\subsubsection{Results}

The results of the comparison between the best selected classifier, XGBoost, with DL and AutoML are presented in in table~\ref{tab:tableDL}. They show that XGBoost over-perform a deep neural network tuned by hand (DNN) F-Score by nearly one point. Comparison between DNN and its optimisation with AutoKeras NAS technique shows an improvement in precision but a loss of recall.The final F-Score of DNN not tuned with NAS over-perform the AutoKeras NAS configuration found with a 4-hour search time. Once the hyperparameters are set, the computational cost of DNN and AutoKeras is in both cases more expensive than the cost of XGBoost calculations. 

\begin{table}
 \caption{Gradient Boosted Trees compared to DNN and Auto-Keras applied on 12 month forecast, 60 days past due, DS1 Train corpus}
  \centering
  \begin{tabular}{llllll}            \\
    \toprule
    \cmidrule(r){1-2}
    Model     & Precision  & Recall & F-Score & AUC & Run time (s) \\
    \midrule
    XGBoost & 0.5138  & 0.5907  & 0.5496 & 0.7822 & 299.8 \\
    DNN & 0.5001 & 0.5865 &  0.5399 & 0.7794 & 450.1     \\
    Auto-Keras Neural Network & 0.5146 & 0.5659 &  0.5390 & 0.7704 & 611      \\
    \bottomrule
  \end{tabular}
  \label{tab:tableDL}
\end{table}

\subsection{Discussion}

In this experiment, the AutoML algorithm with a 4-hour search time did not find a neural layer organisation that would outperform the decision tree algorithms. There is still need in our specific application context, of carefully engineered features used to train decision tree family of algorithms and obtain the best classification performances. Even in the case of hybrid solution involving fusion or ensemble methods to merge outputs from decision trees and DL models, there still is a need for an engineering step to construct the final classification system. Indeed, the assumption frequently claimed to promote NAS technique that, using AutoML algorithms, an operator with minimal machine learning expertise can obtain state-of-the-art performance with very little effort proved to be refuted in our experimental context. 

%\subsubsection{The cost / value ratio of AutoML techniques}

We also note that a new field of research emerge to investigate the energy cost of computation generated by AutoML techniques. In their paper, authors of \cite{strubell2019energy} estimate the cost of AutoML and NAS algorithms for state of the art recent NLP applications. They show that computational costs - and associated carbon impact implied by energy consumption - of modelling using NAS technique on applications like BERT or some transformers, increased along the last 5 years. The authors suggest that it would be beneficial to directly compare different models to perform a cost-benefit analysis. The comparison conducted in this paper could be easily adapted to estimate the cost / value ratio of various algorithms in an application like ours.  

\section{Conclusions and perspectives}
 \label{sec:conclusions}
%% rewrite
The results from this empirical study indicate that the random forest and gradient boosting classifiers obtain the best performances in a user credit card behavior forecasting experiment and are able to cope comparatively well with pronounced class imbalances. Those results are consistent with previously published work on similar topic. We have shown that deep learning models directly applied, using the same features as those selected for decision tree models do not outperform decision tree approaches on our highly unbalanced data set. We also found that the AutoML techniques like NAS, used to automatically configure the neural architecture and hyperparameters, instead of performing well like it have been shown for example with image recognition data sets, slightly underperforms when applied on highly unbalanced data set and small sized corpus like the one used in this study.  

\subsection{Perspectives}

However, since the AutoML method has been shown to increase substantially the precision of the classifier while reducing its recall, we would like to investigate, as potential future work, how the fusion in an hybrid system of the decision tree and DL classifiers optimized with NAS could improve the overall results of classification. 

\bibliographystyle{unsrt}  
\bibliography{references}  %%% Remove comment to use the external .bib file (using bibtex).
%%% and comment out the ``thebibliography'' section.

\end{document}